\documentclass[sigconf]{acmart}
\AtBeginDocument{%
  }

\usepackage{algorithm}
\usepackage{algorithmic}

\usepackage{amsmath}
\usepackage{amsthm}
\usepackage{pifont}

\usepackage{booktabs}
\usepackage{multirow}
\usepackage{rotating}
\usepackage{balance}

\usepackage{array}
\newcommand{\PreserveBackslash}[1]{\let\temp=\\#1\let\\=\temp}
\newcolumntype{C}[1]{>{\PreserveBackslash\centering}p{#1}}
\newcolumntype{R}[1]{>{\PreserveBackslash\raggedleft}p{#1}}
\newcolumntype{L}[1]{>{\PreserveBackslash\raggedright}p{#1}}

\def\eqref#1{equation~\ref{#1}}

\def\1{\bm{1}}

\DeclareMathAlphabet{\mathsfit}{\encodingdefault}{\sfdefault}{m}{sl}
\SetMathAlphabet{\mathsfit}{bold}{\encodingdefault}{\sfdefault}{bx}{n}

\usepackage{tikz}
\usetikzlibrary{arrows.meta, positioning, fit, backgrounds, calc}
\usepackage{xcolor}

\copyrightyear{2026}
\acmYear{2026}
\setcopyright{cc}
\setcctype{by}
\acmConference[KDD '26]{Proceedings of the 32nd ACM SIGKDD Conference on Knowledge Discovery and Data Mining V.2}{August 09--13, 2026}{Jeju Island, Republic of Korea}
\acmBooktitle{Proceedings of the 32nd ACM SIGKDD Conference on Knowledge Discovery and Data Mining V.2 (KDD '26), August 09--13, 2026, Jeju Island, Republic of Korea}
\acmDOI{10.1145/3770855.3818661}
\acmISBN{979-8-4007-2259-2/2026/08}

\begin{document}

\title{Silent Failures in Federated Personalization of Foundation Models}

\author{YongKyung Oh}
\affiliation{%
  \institution{Medical \& Imaging Informatics (MII) Group, \\ 
  University of California, Los Angeles (UCLA)}
  \city{Los Angeles, CA, 90024}
  \country{USA}}
\email{yongkyungoh@mednet.ucla.edu}

\author{Alex Bui}
\affiliation{%
  \institution{Medical \& Imaging Informatics (MII) Group, \\ 
  University of California, Los Angeles (UCLA)}
  \city{Los Angeles, CA, 90024}
  \country{USA}}
\email{buia@mii.ucla.edu}


\begin{abstract}
    Foundation models are increasingly personalized on decentralized private data through federated learning and are now deployed at scale under growing regulatory requirements for post-market monitoring. We argue that this convergence creates a distinct and under-recognized class of trustworthiness failures, which we term ``Silent Failures.'' These include amplified bias, fairness collapse, and alignment erosion that may remain difficult to detect because federated learning's privacy constraints limit visibility into model behavior. A landscape analysis of existing benchmarks reveals a structural divide. Federated benchmarks evaluate system performance but provide limited insight into model behavior, whereas centralized trustworthiness benchmarks assess behavior but require model access incompatible with federated privacy. We introduce a taxonomy of six silent failure modes arising from the interaction of foundation model personalization, dataset shift, and core federated constraints. Our analysis shows that privacy-preserving training alone is insufficient for trustworthy deployment. We conclude with a research agenda for privacy-preserving behavioral evaluation and propose that silent failures become a standard diagnostic category for trustworthy federated artificial intelligence.
\end{abstract}




\maketitle

\section{Introduction}\label{sec:intro}
Foundation models (FMs) are increasingly personalized on decentralized private data through federated learning (FL)~\citep{ren_advances_2025,qiao_towards_2025}, a paradigm we refer to as \textbf{Federated Foundation Model Personalization} (FedFMP).
Parameter-efficient fine-tuning methods such as Low-Rank Adaptation (LoRA)~\citep{hu_lora_2022} enable on-device adaptation of large models, while FL preserves privacy by keeping training data local~\citep{li_federated_2020}. 
Recent methods support heterogeneous devices~\citep{cho_heterogeneous_2024}, federated instruction tuning~\citep{ye_openfedllm_2024,ye_fedllm-bench_2024}, and cross-domain fine-tuning~\citep{gao_flowertune_2025}.

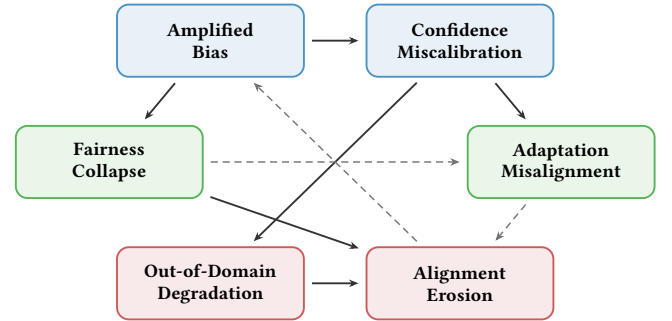
\begin{figure}[htb]
\centering\captionsetup{skip=5pt}
\resizebox{\columnwidth}{!}{%
\begin{tikzpicture}[
  node distance=1.0cm and 0.5cm,
  every node/.style={font=\small},
  box/.style={
    rectangle, rounded corners=5pt,
    minimum width=2.7cm, minimum height=1.0cm,
    align=center, inner sep=1pt
  },
  dir1/.style={draw=dir1col!80, thick, fill=dir1col!10},
  dir2/.style={draw=dir2col!80, thick, fill=dir2col!10},
  dir3/.style={draw=dir3col!80, thick, fill=dir3col!10},
  primary/.style={-{Stealth[length=4pt,width=3.2pt]},
    thick, black!80, shorten <=3pt, shorten >=3pt},
  secondary/.style={-{Stealth[length=4pt,width=3.2pt]},
    semithick, black!55, densely dashed,
    shorten <=3pt, shorten >=3pt}
]

\definecolor{dir1col}{RGB}{31,119,180}
\definecolor{dir2col}{RGB}{44,160,44}
\definecolor{dir3col}{RGB}{190,50,50}

\newcommand{\failurelabel}[2]{%
  \begin{minipage}[c][1.0cm][c]{2.7cm}%
    \centering
    \textbf{#1}\\[-0.25ex]
    \textbf{#2}%
  \end{minipage}%
}

\node[box, dir1] (bias)  {\failurelabel{Amplified}{Bias}};
\node[box, dir1] (calib) [right=0.9cm of bias]  {\failurelabel{Confidence}{Miscalibration}};

\node[box, dir2] (fair)  [below=0.7cm of bias,  xshift=-1.5cm] {\failurelabel{Fairness}{Collapse}};
\node[box, dir2] (adapt) [below=0.7cm of calib, xshift= 1.5cm] {\failurelabel{Adaptation}{Misalignment}};

\node[box, dir3] (ood)   [below=0.7cm of fair,  xshift= 1.5cm] {\failurelabel{Out-of-Domain}{Degradation}};
\node[box, dir3] (align) [below=0.7cm of adapt, xshift=-1.5cm] {\failurelabel{Alignment}{Erosion}};

\draw[primary]   (bias)  -- (fair);
\draw[primary]   (calib) -- (adapt);
\draw[primary]   (ood)   -- (align);
\draw[primary]   (bias)  -- (calib);
\draw[primary]   (calib) -- (ood);
\draw[primary]   (fair)  -- (align);
\draw[secondary] (fair)  -- (adapt);
\draw[secondary] (adapt) -- (align);
\draw[secondary] (align) -- (bias);

\end{tikzpicture}%
}
\caption{%
  Taxonomy of six ``Silent Failures'' in FedFMP, grouped by causal layer:
  \textcolor[RGB]{31,119,180}{\textbf{Data-level}},
  \textcolor[RGB]{44,160,44}{\textbf{Model-level}}, and
  \textcolor[RGB]{190,50,50}{\textbf{System-level}}.
  (Solid arrows denote primary conceptual links, and dashed arrows denote indirect feedback pathways. Arrows indicate conceptual influence rather than strict guarantees.)
}
\label{fig:silent_failures}
\Description{A six-node taxonomy diagram showing amplified bias and confidence miscalibration at the data level, fairness collapse and adaptation misalignment at the model level, and out-of-domain degradation and alignment erosion at the system level, with arrows indicating conceptual influence among the failure modes.}
\end{figure}

Federated foundation models remain hard to audit behaviorally. The privacy constraints that make FL valuable -- secure aggregation, limited visibility, and heterogeneous clients --  limit the behavioral monitoring required for trustworthy deployment~\citep{kairouz_advances_2021,lyu_privacy_2024}.
Our contribution is to identify silent failures as a distinct failure category in FedFMP rather than as a single new failure mode.
We refer to the failures that arise under these conditions as ``Silent Failures'' as summarized in Figure~\ref{fig:silent_failures} and Table~\ref{tab:failure_modes_summary}. These failures manifest as degradations in model behavior that remain difficult to detect as FL restricts access to training data and intermediate model signals. We question treating privacy-preserving training as sufficient. 

\begin{table*}[htb]
\footnotesize\centering\captionsetup{skip=5pt}
\caption{Taxonomy of six silent failures in FedFMP, organized by causal layer. 
}
\label{tab:failure_modes_summary}
\begin{tabular}{@{}c c m{5.6cm} m{4.2cm} m{4.0cm}@{}}
\toprule
\textbf{~} & \textbf{Failure Mode} & \multicolumn{1}{c}{\textbf{Description}} & \multicolumn{1}{c}{\textbf{FM Perspective}} & \multicolumn{1}{c}{\textbf{Dataset Shift Factor}} \\
\midrule
\multirow{5}{*}{\rotatebox{90}{\textbf{Data-level}}}
& \begin{tabular}[c]{@{}c@{}}Amplified \\ Bias\end{tabular}
& Personalization on skewed local data can amplify biases; privacy constraints hinder centralized auditing of local disparities.
& Local objective differences or biased prompts can exacerbate output disparities during adaptation.
& Within-client subpopulation shifts and skewed label distributions. \\
\cmidrule{2-5}
& \begin{tabular}[c]{@{}c@{}}Confidence \\ Miscalibration\end{tabular}
& Model confidence can decouple from accuracy on shifted data; global aggregation may mask localized overconfident errors.
& Local fine-tuning may reinforce hallucination patterns, producing confidently incorrect generations.
& Covariate or concept shifts that alter local data factuality. \\
\midrule
\multirow{5}{*}{\rotatebox{90}{\textbf{Model-level}}}
& \begin{tabular}[c]{@{}c@{}}Fairness \\ Collapse\end{tabular}
& Aggregating adapters tuned to distinct distributions can widen disparities. Affected subgroups may be too small to alter global metrics.
& Personalization can overfit to prevalent local prompts, exacerbating the FM's inherent prompt sensitivity.
& Between-client divergence in prompt-style distributions across heterogeneous user populations. \\
\cmidrule{2-5}
& \begin{tabular}[c]{@{}c@{}}Adaptation \\ Misalignment\end{tabular}
& Overfitting to narrow local objectives may degrade general capabilities and safety. The server may not distinguish adaptation from forgetting.
& In PEFT-based personalization, small adapter updates can alter safety behavior encoded in the frozen backbone.
& Deviations in local task objectives or rapid concept drift. \\
\midrule
\multirow{5}{*}{\rotatebox{90}{\textbf{System-level}}}
& \begin{tabular}[c]{@{}c@{}}Out-of-Domain \\ Degradation\end{tabular}
& Performance can degrade outside the local specialization domain. Privacy limits cross-client OOD evaluation.
& Over-specialization on narrow local data weakens the FM's zero-shot generalization capabilities.
& Sustained covariate shift narrowing the effective training distribution to a specialized local domain. \\
\cmidrule{2-5}
& \begin{tabular}[c]{@{}c@{}}Alignment \\ Erosion\end{tabular}
& Local degradation of behaviors (safety, persona) may be masked by aggregation. The server lacks local-output access for audits.
& Shifting local content may weaken safety guardrails and increase hallucination propensity.
& Concept drift in local content that shifts the behavioral ground-truth. \\
\bottomrule
\end{tabular}
\vspace{-5pt}
\end{table*}


Three trends motivate this focus.
(i)~\textbf{Deployment scale}. Federated personalization is moving toward large client populations, where rare failures can still affect many users.
(ii)~\textbf{Empirical risk}. Recent studies show that fine-tuning on as few as ten examples can compromise safety alignment~\citep{qi_fine-tuning_2024}. Benchmark evaluations may also be corrupted by data contamination even in centralized settings~\citep{balloccu_leak_2024,li_task_2024}. These risks increase when training occurs across heterogeneous and partially observable client environments.
(iii)~\textbf{Accountability}. As artificial intelligence (AI) systems are deployed in high-stakes domains such as healthcare and finance, emerging governance frameworks increasingly require continuous post-deployment monitoring of model behavior~\citep{bommasani_opportunities_2021}. Current federated evaluation tools provide limited support for such monitoring.

This paper argues that \emph{silent failures form a distinct problem category}. 
Their defining property is that the privacy architecture of FL reduces the observability needed for detection, not only for mitigation. We show this gap through an analysis of twelve representative benchmarks spanning federated and centralized evaluation settings. 
Addressing these failures requires evaluation tools that operate under realistic privacy constraints and heterogeneous client conditions, which current frameworks only partly support.

\section{Problem Formulation}\label{sec:problem}
Consider a federated network of $K$ clients coordinated by a central server. 
Each client $k$ personalizes a pre-trained FM $\mathcal{F}$ with frozen parameters $\theta_{\text{pre}}$ by training a small adapter $\Delta\theta_k$ via PEFT, producing outputs $\mathcal{F}(x;\theta_{\text{pre}},\Delta\theta_k)$.
Then, client $k$ holds a private dataset $D_k^{(t)}$ drawn from a local distribution $P_k^{(t)}(X,Y)$ that may shift over time. In each round, the client minimizes a local objective
\begin{align*}
\Delta\theta_k^{(t+1)} \leftarrow 
\arg\min_{\Delta\theta}
\mathbb{E}_{(x,y)\sim D_k^{(t)}}
[\mathcal{L}(\mathcal{F}(x;\theta_{\text{pre}},\Delta\theta),y)] .
\end{align*}

The server aggregates updates via FedAvg
\begin{align*}
\Delta\theta_{\text{agg}}^{(t+1)} =
\frac{1}{K}\sum_{k=1}^{K}\Delta\theta_k^{(t+1)} .
\end{align*}
We adopt this simple aggregation rule as a baseline for analyzing failure modes rather than as an optimal strategy.

The central challenge is \textit{dataset shift}~\citep{moreno-torres_unifying_2012,oh_multi-view_2025}. Distributions may vary across clients ($P_k \neq P_j$) and drift within a client over time ($P_k^{(t)} \neq P_k^{(t-1)}$).
Because FL limits the server's local-data view, these shifts can be hard to observe. This information asymmetry, combined with the frozen backbone's limited adaptability and simple aggregation, creates conditions for silent failures, where trustworthiness can degrade without triggering standard performance alarms.

\section{Silent Failures from FedFMP}\label{sec:failure}
We identify six failure modes organized by causal layer in Figure~\ref{fig:silent_failures} and Table~\ref{tab:failure_modes_summary}. These failures span three causal layers, each with a distinct origin:
\begin{itemize}
    \item Data-level failures (amplified bias, confidence miscalibration) originate from skewed or noisy local inputs that can distort adapter updates.
    \item Model-level failures (fairness collapse, adaptation misalignment) arise when aggregation combines adapters shaped by conflicting local objectives.
    \item System-level failures (Out-of-Domain degradation, alignment erosion) appear as behavioral drift after deployment and accumulate across training rounds.
\end{itemize}

In all cases, FL privacy constraints reduce the observability needed for detection. Table~\ref{tab:failure_modes_summary} characterizes each failure from the perspectives of foundation models and dataset shift. Table~\ref{tab:fl_constraints_influence} shows how three core FL constraints, aggregation, limited data observability, and statistical heterogeneity, amplify or conceal each mode.

\begin{table*}[htb]
\footnotesize\centering\captionsetup{skip=5pt}
\caption{Influence of core FL constraints on each silent failure mode.}
\label{tab:fl_constraints_influence}
\begin{tabular}{@{}c m{5.0cm} m{4.2cm} m{5.4cm}@{}}
\toprule
\textbf{Failure Mode}
  & \multicolumn{1}{c}{\textbf{Aggregation}}
  & \multicolumn{1}{c}{\textbf{Limited Data Observability}}
  & \multicolumn{1}{c}{\textbf{Statistical Heterogeneity}} \\
\midrule
\begin{tabular}[c]{@{}c@{}}Amplified \\ Bias\end{tabular}
& Biased local updates can compound when divergent demographic skews reinforce rather than cancel.
& Local demographic and label distributions are hard to audit centrally.
& Skewed local demographics or labels can lead adapters to encode different bias directions. \\
\midrule
\begin{tabular}[c]{@{}c@{}}Confidence \\ Miscalibration\end{tabular}
& Averaging locally overconfident adapters can smooth elevated ECE signals, concealing domain-specific miscalibration in the global model.
& Local output factuality is hard to verify; global validation may lack domain specificity.
& Varying data factuality and domain coverage across clients can produce heterogeneous calibration errors. \\
\midrule
\begin{tabular}[c]{@{}c@{}}Fairness \\ Collapse\end{tabular}
& Averaging adapters overfitted to divergent prompt styles can produce an uneven performance profile across clients.
& Local prompt distributions and per-client subgroup performance are not visible to the central server.
& Between-client divergence can specialize adapters for different populations, producing uneven behavior. \\
\midrule
\begin{tabular}[c]{@{}c@{}}Adaptation \\ Misalignment\end{tabular}
& Aggregating adapters that alter different safety-relevant subspaces can produce global incoherence within a round.
& Local alignment decay is hidden from the server, allowing safety-relevant forgetting with limited detection.
& Diverse local tasks can make adapters diverge, amplifying incoherence upon aggregation. \\
\midrule
\begin{tabular}[c]{@{}c@{}}Out-of-Domain \\ Degradation\end{tabular}
& Averaging over-specialized adapters can weaken cross-client generalization.
& The server may not diagnose over-specialization, so OOD fragility can accumulate until deployment.
& High heterogeneity can make each client's data effectively OOD for others; FM specialization amplifies this fragility. \\
\midrule
\begin{tabular}[c]{@{}c@{}}Alignment \\ Erosion\end{tabular}
& Per-round accumulation of small behavioral degradations can compound before detection.
& Without local outputs, per-client compliance scores are uncomputable at the server, so cumulative erosion can be missed.
& Clients may drift toward different behavioral standards as their local content evolves, making global alignment a moving target. \\
\bottomrule
\end{tabular}
\vspace{-5pt}
\end{table*}


\subsection{Amplified Bias}\label{sec:bias}

In FedFMP, bias can be amplified when clients personalize on skewed local data with conflicting objectives~\citep{chang_bias_2023}. Let $M(\mathcal{F}, s)$ denote a model-performance metric evaluated on a data slice $s$, where $s$ may represent a demographic group or a client distribution. Examples include accuracy, calibration, and generation-bias measures such as toxicity, regard, and gender polarity~\citep{dhamala_bold_2021}.
Following standard group-fairness formulations~\citep{dwork_fairness_2012}, we quantify bias as the disparity of $M$ between two demographic groups $g_a$ and $g_b$:
\begin{align*}
\text{Bias}(\mathcal{F}) = |M(\mathcal{F}, g_a) - M(\mathcal{F}, g_b)|.
\end{align*}

Amplification occurs when the aggregated model exhibits greater bias than the pre-trained baseline, i.e., $\text{Bias}(\mathcal{F}_{\text{agg}}) > \text{Bias}(\mathcal{F}_{\text{pre}})$. Although averaging may seem to reduce bias, clients with biases in different directions produce conflicting updates whose aggregation can yield interference and global bias beyond any individual client's~\citep{wyllie_fairness_2024,mcmahan_communication-efficient_2017}. FL privacy limits auditability, so amplification may be hard to detect.
As shown in Figure~\ref{fig:silent_failures}, amplified bias creates conditions that increase the risk of confidence miscalibration (Section~\ref{sec:miscalib}) and fairness collapse (Section~\ref{sec:fairness}).

\subsection{Confidence Miscalibration}\label{sec:miscalib}

Foundation models can produce fluent but incorrect content~\citep{ji_survey_2023}. Under distribution shift, calibration studies show high-confidence errors and elevated ECE~\citep{oh_towards_2024}. In FedFMP, local factual noise may reinforce such errors during personalization. A common measure of miscalibration is the Expected Calibration Error (ECE), following calibration literature~\citep{pakdaman_naeini_obtaining_2015,guo_calibration_2017,lee_continuum_2026} and recent confidence-aware calibration work~\citep{zhao_confidence-aware_2024}:
\begin{align*}
\text{ECE} = \sum_{m=1}^{M} \frac{|B_m|}{n} |\text{acc}(B_m) - \text{conf}(B_m)|.
\end{align*}

If a local distribution shift in $P_k^{(t)}$ strengthens hallucinated patterns, the client adapter $\Delta\theta_k$ can increase the local ECE. Aggregation of such adapters may then produce a global model $\mathcal{F}_{\text{agg}}$ with high ECE. 
This failure is harder to detect in federated settings than in centralized ones. In centralized fine-tuning, domain-specific holdout sets can reveal elevated ECE before deployment. In FedFMP, privacy constraints limit inspection of client outputs, and the server often lacks domain-matched validation data to detect localized miscalibration~\citep{kairouz_advances_2021,lyu_privacy_2024}. The ECE signal exists at the client level but may be obscured at the server, where aggregation smooths it out.

Miscalibrated confidence can feed back into adaptation misalignment, which is examined in Section~\ref{sec:misalign}. When local models report unreliable confidence estimates, these signals bias the feedback used for personalization. The resulting updates distort the optimization trajectory and gradually push the system away from the intended objective.

\subsection{Fairness Collapse}\label{sec:fairness}

Unlike bias amplification, which reflects demographic disparities within local data (Table~\ref{tab:failure_modes_summary}), \textit{Fairness Collapse} describes growing performance imbalance across clients caused by heterogeneous personalization. Prompt-based models are sensitive to prompt order and semantics~\citep{lu_fantastically_2022,webson_prompt-based_2022}. In FedFMP, local fine-tuning can adapt to prompt styles prevalent in majority user groups, creating uneven performance across clients.

Aggregation further combines adapters specialized for different local distributions~\citep{li_federated_2020,ye_heterogeneous_2023,pillutla_federated_2022}. As FL restricts access to client data and outputs, the server may not observe local prompt or subgroup patterns, making the resulting imbalance difficult to detect~\citep{chen_privacy_2023}.

We model this collapse through the variance of $M(\mathcal{F}, P_k^{(t)})$ across the $K$ clients, where $M$ is the performance metric defined in Section~\ref{sec:bias} with the client distribution $P_k^{(t)}$ as the data slice. The disparity across clients is measured by the standard deviation
\begin{align*}
\text{Disp}(\mathcal{F}) =
\text{std}\bigl(
M(\mathcal{F}, P_1^{(t)}), \dots, M(\mathcal{F}, P_K^{(t)})
\bigr).
\end{align*}

We identify fairness collapse when $\text{Disp}(\mathcal{F}_{\text{agg}}) \gg \text{Disp}(\mathcal{F}_{\text{pre}})$ beyond legitimate cross-client distributional differences, indicating that aggregation has amplified rather than balanced performance disparities.
This divergence arises because each adapter is optimized for a local distribution with distinct characteristics, and their aggregation amplifies cross-client disparities~\citep{mcmahan_communication-efficient_2017,li_federated_2020}.

This failure differs from the within-dataset demographic bias in Section~\ref{sec:bias}. When clients represent distinct user populations, a global model may underserve some clients relative to others. This pattern creates uneven service quality across the federation. The fairness concern arises from structural heterogeneity between clients, not demographic attributes within a single dataset. Such disparity can contribute to OOD degradation in Section~\ref{sec:ood} when inter-client differences compound distributional mismatch.

\subsection{Adaptation Misalignment}\label{sec:misalign}

Adapting foundation models through federated PEFT introduces a risk that personalization either erodes general capabilities or overfits to narrow local objectives~\citep{zhang_memory_2022,de_lange_continual_2022}. In FedFMP, strong local adaptation signals can cause personalized models to diverge from their base alignment. When such adapters are aggregated across heterogeneous clients, the resulting global model may combine incompatible adaptations. Because FL privacy constraints obscure the internal state of local models, the server may struggle to distinguish beneficial adaptation from safety-relevant forgetting.

This failure can be formalized as a trade-off between a local personalization loss $\mathcal{L}_{\text{local}}$ and a base alignment loss $\mathcal{L}_{\text{base}}$~\citep{de_lange_continual_2022}. Misalignment arises when optimizing the adapter update $\Delta\theta_k$ for the local objective degrades the base alignment:
\begin{align*}
&\text{if } \Delta\theta_k = \arg\min_{\Delta\theta}\,
\mathcal{L}_{\text{local}}\!\left(\mathcal{F}(x;\theta_{\text{pre}},\Delta\theta)\right), \\
&\text{then } 
\mathcal{L}_{\text{base}}(\mathcal{F}_k) \gg 
\mathcal{L}_{\text{base}}(\mathcal{F}_{\text{pre}}),
\end{align*}
where $\mathcal{F}_k$ denotes the personalized model. Although adapter parameters are compact ($|\Delta\theta_k| \ll |\theta_{\text{pre}}|$), PEFT updates concentrate on high-impact parameter subspaces (Table~\ref{tab:failure_modes_summary}). 
Small adapters can alter safety-critical behavior encoded in the frozen backbone.
This vulnerability appears even in centralized fine-tuning settings~\citep{qi_fine-tuning_2024} and is relevant to alignment objectives studied in RLHF~\citep{bai_training_2022}.
Aggregating adapters that forget different aspects of the base alignment may therefore produce a globally inconsistent model~\citep{mcmahan_communication-efficient_2017,li_federated_2020}. 

Misaligned adapters can exacerbate OOD degradation in Section~\ref{sec:ood} and feed back into alignment erosion in Section~\ref{sec:erosion} over successive rounds.

\subsection{Out-of-Domain Degradation}\label{sec:ood}

In FedFMP, intensive personalization can cause a model to overfit to a client's narrow local data distribution. The resulting model performs well locally but degrades on data from other clients~\citep{hendrycks_many_2021,sagawa_distributionally_2020}. Because FL environments are highly heterogeneous, clients may effectively act as out-of-domain environments for each other. At the same time, privacy constraints limit server diagnosis of the over-specialization that leads to this failure.

The failure appears when a personalized model $\mathcal{F}_k$, trained on client $k$'s distribution $P_k^{(t)}$, is evaluated on the distribution of a different client $j$:
\begin{align*}
M(\mathcal{F}_k, P_j^{(t)}) \ll M(\mathcal{F}_k, P_k^{(t)}).
\end{align*}
This behavior reflects a well-known challenge in heterogeneous FL~\citep{ye_heterogeneous_2023}. When such over-specialized adapters are aggregated, the resulting global model $\mathcal{F}_{\text{agg}}$ can exhibit poor generalization. This limitation motivates research on shift-robust personalization~\citep{zhang_dm-pfl_2023}, and federated OOD learning frameworks~\citep{liao_foogd_2024}.

In FedFMP, it can also erode the zero-shot generalization capability of the foundation model, creating a trade-off that does not arise in task-specific models.
OOD degradation marks a convergence point in the taxonomy. It can arise from fairness collapse and adaptation misalignment and contribute to alignment erosion.

\subsection{Alignment Erosion}\label{sec:erosion}

Alignment erosion refers to the gradual degradation of a model's safety alignment, persona, or factual reliability even when task accuracy remains high~\citep{ganguli_predictability_2022,bai_training_2022}. In FedFMP, this occurs when local personalization weakens safety guardrails due to shifts in local content or evaluation criteria~\citep{yang_impact_2023,qi_fine-tuning_2024}. Aggregation can mask these small degradations, while FL privacy constraints limit inspection of local outputs that might reveal behavioral violations.

We model this phenomenon as concept drift~\citep{lu_learning_2019, bayram_concept_2022, li_concept_2024} in adherence to a set of behavioral constraints $\mathcal{C}=\{c_1,c_2,\dots\}$. Let $S(\mathcal{F},c)\in[0,1]$ denote a score that measures compliance with constraint $c$, computed using a reward model trained via RLHF~\citep{bai_training_2022}. Motivated by hallucination and safety-degradation evidence~\citep{ji_survey_2023,qi_fine-tuning_2024}, erosion occurs when optimizing a local objective $\mathcal{L}_k^{(t)}$ on a shifted distribution $P_k^{(t)}$ produces a personalized model $\mathcal{F}_k$ with reduced compliance, $S(\mathcal{F}_k,c) < S(\mathcal{F}_{\text{pre}},c)$. Because the server may not compute $S(\mathcal{F}_k,c)$ for each client, aggregation of these slightly degraded adapters may proceed with limited detection. 
Because each round's aggregated model $\mathcal{F}_{\text{agg}}^{(t)}$ initializes the next local training round, degradations may compound: over $T$ rounds, even small decrements $\delta S$ can accumulate so that $S(\mathcal{F}_{\text{agg}}^{(T)},c) \ll S(\mathcal{F}_{\text{pre}},c)$~\citep{kairouz_advances_2021}, enabled by the aggregation and observability constraints in Table~\ref{tab:fl_constraints_influence}.

Unlike adaptation misalignment (Section~\ref{sec:misalign}), which emerges from a single round of local optimization, alignment erosion can accumulate across many aggregation rounds, fed by upstream OOD degradation and misaligned adapters.

\section{The Monitoring Gap in Federated Systems}\label{sec:gap}
\begin{table*}[t]
\footnotesize\centering\captionsetup{skip=5pt}
\caption{Benchmark Landscape Analysis
(\ding{51}\,:\,Direct Measurement,\,
 $\triangle$\,:\,Proxy/Partial,\,
 --\,:\,Not in Scope.)}
\label{tab:benchmark_landscape}
\begin{tabular}{@{}l l l C{1.6cm}C{1.6cm}C{1.6cm}C{1.6cm}C{1.6cm}C{1.6cm}@{}}
\toprule
~ & \textbf{Benchmark Name} & \textbf{Primary Focus} &
\begin{tabular}[c]{@{}c@{}}Amplified \\ Bias\end{tabular} &
\begin{tabular}[c]{@{}c@{}}Confidence \\ Miscalibration\end{tabular} &
\begin{tabular}[c]{@{}c@{}}Fairness \\ Collapse\end{tabular} &
\begin{tabular}[c]{@{}c@{}}Adaptation \\ Misalignment\end{tabular} &
\begin{tabular}[c]{@{}c@{}}Out-of-Domain \\ Degradation\end{tabular} &
\begin{tabular}[c]{@{}c@{}}Alignment \\ Erosion\end{tabular} \\
\midrule
\multirow{6}{*}{\rotatebox{90}{\textbf{Federated}}}
& LEAF~\cite{caldas_leaf_2018}               & FL Heterogeneity   & $\triangle$ & --          & \ding{51}   & --          & $\triangle$ & --          \\
& FedScale~\cite{lai_fedscale_2022}           & FL Systems          & --          & --          & $\triangle$ & --          & $\triangle$ & --          \\
& FLamby~\cite{ogier_du_terrail_flamby_2022} & FL Healthcare       & $\triangle$ & --          & $\triangle$ & $\triangle$ & $\triangle$ & --          \\
& FedLLM-Bench~\cite{ye_fedllm-bench_2024}   & FL LLM Tuning       & $\triangle$ & --          & $\triangle$ & \ding{51}   & $\triangle$ & $\triangle$ \\
& FLHetBench~\cite{zhang_flhetbench_2024}    & FL Systems          & --          & --          & --          & --          & $\triangle$ & --          \\
& PFLlib~\cite{zhang_pfllib_2025}            & FL Personalization  & --          & --          & --          & --          & --          & --          \\
\midrule
\multirow{6}{*}{\rotatebox{90}{\textbf{Centralized}}}
& HELM~\cite{liang_holistic_2023}             & Holistic Evaluation & \ding{51}   & \ding{51}   & --          & --          & $\triangle$ & $\triangle$ \\
& BIG-bench~\cite{srivastava_beyond_2023}     & Reasoning           & $\triangle$ & \ding{51}   & --          & --          & $\triangle$ & --          \\
& DecodingTrust~\cite{wang_decodingtrust_2023}& LLM Trust           & \ding{51}   & --          & $\triangle$ & --          & \ding{51}   & \ding{51}   \\
& JailbreakBench~\cite{chao_jailbreakbench_2024}& Safety            & --          & --          & --          & --          & --          & \ding{51}   \\
& TrustLLM~\cite{huang_position_2024}         & Trust Evaluation    & \ding{51}   & --          & --          & --          & \ding{51}   & \ding{51}   \\
& HarmBench~\cite{mazeika_harmbench_2024}     & Red Teaming         & --          & --          & --          & --          & --          & \ding{51}   \\
\bottomrule
\end{tabular}
\vspace{-5pt}
\end{table*}

FedFMP still lacks mature behavioral monitoring. This section explains why. We examine the conflict between FL privacy and behavioral evaluation, the mechanisms that mask failures, and the evidence from existing benchmarks.

\paragraph{The Foundational Conflict from FL Constraints.}
The monitoring gap arises from a structural difference between centralized and federated systems. 
In centralized settings, monitoring relies on direct access to data, model internals, and ground-truth labels~\citep{lu_learning_2019, bayram_concept_2022, li_concept_2024}. 
Even with this access, reliable evaluation remains difficult. Benchmark data can leak into model training~\citep{balloccu_leak_2024}. Reported performance may reflect contamination rather than generalization~\citep{li_task_2024}. Detecting contamination without access to training data remains difficult~\citep{golchin_time_2024}.
In contrast, federated learning enforces privacy-by-design constraints that restrict such inspection~\citep{kairouz_advances_2021}. As a result, many established monitoring methods become difficult to apply~\citep{lyu_privacy_2024}. This difference in observability has produced two distinct evaluation paradigms that are difficult to reconcile.

\paragraph{Systemic Mechanisms of Detection Failure.}
The interaction between FM behavioral failures (Table~\ref{tab:failure_modes_summary}) and FL constraints (Table~\ref{tab:fl_constraints_influence}) can conceal silent failures. Two effects are central. First, privacy introduces a trade-off with monitoring resolution. Secure aggregation protects client data by hiding individual updates~\citep{mansouri_learning_2022, lyu_privacy_2024}, but this aggregation masks local signals that often reveal emerging failures~\citep{kairouz_advances_2021, zhang_survey_2023}. Second, reduced visibility creates statistical masking. Failures that affect a small subset of clients may not appear in global metrics. For example, fairness collapse in a minority population can remain hidden if its impact on the global average is small~\citep{chen_privacy_2023, wyllie_fairness_2024}. Similar masking can occur for miscalibration or catastrophic forgetting~\citep{oh_towards_2024, zhang_memory_2022}.

\paragraph{Benchmark Landscape Analysis.}
This structural limitation appears in the benchmark ecosystem. To examine the issue empirically, we analyze the current evaluation landscape, summarized in Table~\ref{tab:benchmark_landscape}. Prior surveys~\citep{chang_survey_2024,yao_survey_2024,zhao_survey_2026} mainly catalogue benchmark coverage. Our analysis instead evaluates how benchmarks capture specific failure modes. The results suggest a consistent divide. Federated benchmarks measure system-level performance but offer limited visibility into model behavior. Centralized trustworthiness benchmarks measure behavioral properties but assume direct model access, which conflicts with FL privacy constraints. We find no single framework that spans both paradigms. The monitoring gap therefore reflects a structural limitation rather than a simple oversight.

Each benchmark in Table~\ref{tab:benchmark_landscape} is rated on a three-point scale. A check mark (\ding{51}) indicates direct measurement, meaning the failure mode metric is reported by the benchmark's official evaluation scripts. A triangle ($\triangle$) indicates proxy or partial coverage, where the metric can be derived from standard outputs with limited post-processing or applies only to a subset of tasks. A dash (--) indicates that the benchmark's design does not directly address the failure mode. 
Note that these ratings involve interpretive judgment and may reflect subjective decisions.

\paragraph{Distinction from Adversarial Robustness.}
Silent failures differ from recent research on red teaming~\citep{perez_red_2022, feffer_red-teaming_2024} and jailbreaking~\citep{wei_jailbroken_2023,xu_bag_2024,mai_you_2025}.
Adversarial studies examine failures caused by external attacks. Silent failures arise from internal system dynamics during federated personalization. Adversarial evaluation assumes direct access to the model for probing. Silent failures occur when such access is unavailable because FL privacy constraints restrict observability. This lack of visibility defines the monitoring gap.

\section{Research Vision for Trustworthy FedFMP}\label{sec:agenda}
The monitoring gap identified in Section~\ref{sec:gap} reflects a structural property of FL more than a simple implementation limitation. Privacy constraints restrict the visibility needed to evaluate model behavior. This paper therefore focuses on defining the problem space rather than proposing a single technical fix. 

\paragraph{Robust Aggregation and Personalization.}
Standard aggregation methods optimize for predictive performance but can produce global models whose behavioral properties match no client's local distribution. Robust averaging, architecture-aware grouping~\citep{wang_federated_2020,mansouri_learning_2022,yang_personalized_2023}, and personalization with explicit failure-mode constraints~\citep{pillutla_federated_2022,yang_dual-personalizing_2024} offer partial remedies. 
The central tension is whether an aggregation rule can bound cross-client fairness disparity without sacrificing the local adaptation that makes federated personalization useful in the first place.

\paragraph{Reliable Monitoring and Auditing.}
Detecting behavioral degradation in federated systems requires monitoring methods that operate without direct access to client data or model internals. This is both a tooling gap and a structural consequence of privacy-by-design: secure aggregation conceals the individual updates that most commonly signal emerging failures~\citep{zhang_survey_2023}. Progress requires federated auditing protocols that can assess fairness, calibration, and alignment from aggregate signals alone~\citep{chen_privacy_2023,lyu_privacy_2024}.

\paragraph{Adaptive Personalization and Continual Learning.}
Most personalization strategies assume stable data distributions. In practice, client data evolve, and static adapters can drift behaviorally over time. 
Continual learning mechanisms for mitigating catastrophic forgetting~\citep{zhang_memory_2022,de_lange_continual_2022} offer a starting point. How to integrate shift-aware personalization~\citep{zhang_dm-pfl_2023} with federated continual learning under non-stationary client data, while preserving privacy guarantees, remains an open problem~\citep{fan_ten_2025,ren_advances_2025}.

\paragraph{Rigorous Benchmarking and Explainability.}
Existing frameworks such as OpenFedLLM~\citep{ye_openfedllm_2024} and FlowerTune~\citep{gao_flowertune_2025} provide strong baselines for system-level evaluation. They do not yet treat fairness, calibration, or OOD robustness~\citep{hendrycks_many_2021} under realistic heterogeneous shifts~\citep{sagawa_distributionally_2020,pei_review_2024} as first-class metrics.
A parallel direction studies federated explainability. When direct inspection of client data is unavailable, interpretability applied to aggregated model behavior may partially recover the observability that privacy removes. This approach can turn a constraint into a diagnostic signal.

\paragraph{A Vision for Success.}
If this agenda succeeds, federated AI evaluation will treat behavioral diagnostics as a core concern alongside system performance. Silent failures will become a recognized diagnostic category. Benchmark suites will measure behavioral properties under privacy constraints.
FL frameworks will include privacy-preserving behavioral auditing as a standard capability rather than a bespoke post hoc tool. 
The result will be a federated AI ecosystem where privacy-preserving training is paired with evidence for trustworthy deployment.

\section{Conclusion}\label{sec:conclusion}
Federated personalization of foundation models creates silent failures: behavioral degradations that current evaluation tools often miss. Our taxonomy and twelve-benchmark analysis suggest that this gap is structural. Federated benchmarks track system performance, while centralized trustworthiness benchmarks assume model and data access that FL restricts.

The path forward is not to abandon federated personalization, but to evaluate it differently. Privacy-preserving training should be paired with privacy-preserving checks for fairness collapse, miscalibration, OOD degradation, and alignment erosion. FedFMP systems should therefore be judged not only by whether they train efficiently across clients, but by whether their behavioral failures can be audited under the same privacy constraints that govern training.

\begin{acks}
    Dr. YongKyung Oh was supported by the Basic Science Research Program through the National Research Foundation of Korea (NRF) funded by the Ministry of Education (RS-2024-00407852).
\end{acks}

\bibliographystyle{ACM-Reference-Format}
\balance
\bibliography{references}

\appendix
\section{Benchmark Paper Overviews}\label{app:benchmark-overviews}

Our analysis builds on benchmarks developed by many research groups in federated learning and trustworthiness evaluation. The monitoring gap we identify reflects the independent evolution of these two fields, not a limitation of any single benchmark. We hope this work encourages closer exchange between them.

This appendix summarizes the benchmark papers used in Table~\ref{tab:benchmark_landscape}. Each note describes what the benchmark measures and where its coverage stops for ``Silent Failures'' in \textbf{Federated Foundation Model Personalization} (FedFMP).
Because these ratings involve interpretive judgment, we release the rubric and per-benchmark worksheets to support audit and refinement. For details, please refer to \url{https://github.com/yongkyung-oh/FedFMP}.

\subsection{Federated Benchmarks}

\paragraph{LEAF}
This modular benchmark suite covers federated settings through datasets, protocols, and reference implementations built around statistical and systems heterogeneity~\cite{caldas_leaf_2018}. It captures client imbalance and variation. Behavioral failures such as alignment erosion or confidence miscalibration are outside its scope.

\paragraph{FedScale}
This benchmark combines federated datasets with a scalable runtime for tasks including image classification, object detection, language modeling, and speech recognition~\cite{lai_fedscale_2022}. Its evidence is primarily systems-level. Trustworthiness appears only through proxies such as heterogeneity and robustness.

\paragraph{FLamby}
This benchmark targets cross-silo federated learning in healthcare using natural institutional splits, multiple modalities, and baseline implementations~\cite{ogier_du_terrail_flamby_2022}. These splits expose fairness and out-of-domain risks across clinical sites. Foundation-model alignment and safety behavior lie outside its scope.

\paragraph{FedLLM-Bench}
This benchmark evaluates federated learning for large language models using datasets for instruction tuning and preference alignment, together with multiple training methods and metrics~\cite{ye_fedllm-bench_2024}. It is the only federated benchmark in this set that explicitly covers LLM tuning and preference alignment. The metrics center on training outcomes rather than privacy-preserving behavioral monitoring.

\paragraph{FLHetBench}
This benchmark studies device and state heterogeneity in federated learning through sampling methods and metrics for client availability and resource variation~\cite{zhang_flhetbench_2024}. For silent failures, its role is system-level: it shows how availability and resource constraints can mask problems. It does not test demographic, safety, or alignment degradation.

\paragraph{PFLlib}
This benchmark library covers many FL and pFL algorithms across heterogeneous scenarios and datasets~\cite{zhang_pfllib_2025}. It supports evaluation of personalization and cross-client generalization. The reported outcomes remain algorithmic rather than behavioral.

\subsection{Centralized Trustworthiness Benchmarks}

\paragraph{HELM}
This benchmark evaluates language models across scenarios, metrics, and behaviors including accuracy, calibration, robustness, fairness, bias, and toxicity~\cite{liang_holistic_2023}. Its behavioral coverage is wider than that of the federated benchmarks, although the centralized setup assumes access to inputs and outputs that federated privacy constraints may restrict.

\paragraph{BIG-bench}
This benchmark collects a large set of tasks for measuring and extrapolating language-model capabilities~\cite{srivastava_beyond_2023}. It probes reasoning and capability limits. Federated heterogeneity, personalization, and deployment-time trustworthiness failures are not its target.

\paragraph{DecodingTrust}
This benchmark evaluates GPT-model trustworthiness across toxicity, stereotype bias, adversarial robustness, privacy, ethics, fairness, and out-of-domain robustness~\cite{wang_decodingtrust_2023}. It covers many behavioral risks. It does not model the federated mechanism by which local failures can disappear during aggregation.

\paragraph{JailbreakBench}
This benchmark standardizes evaluation of jailbreak attacks and defenses through open adversarial prompts, a fixed behavior set, scoring functions, and a public leaderboard~\cite{chao_jailbreakbench_2024}. It informs alignment-erosion analysis. Its unit of evaluation is adversarial safety rather than federated personalization or client-level monitoring.

\paragraph{TrustLLM}
This trustworthiness benchmark covers large language models across truthfulness, safety, fairness, robustness, privacy, and ethics~\cite{huang_position_2024}. These categories overlap with silent failures. The evaluation, however, does not model federated privacy constraints.

\paragraph{HarmBench}
This benchmark evaluates automated red teaming and robust refusal across attack methods, target models, and harmful behavior categories~\cite{mazeika_harmbench_2024}. It is relevant to alignment and safety failures. It does not study how those failures arise or stay hidden during federated personalization.


\vfill\eject

\end{document}